\documentclass[5p,twocolumn]{elsarticle}   

\usepackage{amsmath,amssymb,amsfonts}
\usepackage{xcolor}
\usepackage{soul}
\usepackage{float}
\usepackage{caption}
\usepackage{booktabs}
\usepackage{xurl}
\usepackage{pifont}
\usepackage{graphicx}
\usepackage{textcomp}
\usepackage{comment}
\usepackage{listings}
\usepackage[english]{babel}
\usepackage[acronym]{glossaries}
\newacronym{ai}{AI}{Artificial Intelligence}
\newacronym{ml}{ML}{Machine Learning}
\newacronym{edc}{EDC}{Eclipse Dataspace Components}
\newacronym{llm}{LLM}{Large Language Model}
\newacronym{mvds}{MVD}{Minimal Viable Data Space}
\newacronym{mcp}{MCP}{Model Context Protocol}
\newacronym{tui}{TUI}{Terminal User Interface}
\newacronym{ids}{IDS}{
International Data Spaces}
\newacronym{idsa}{IDSA}{
International Data Spaces Association}
\newacronym{iot}{IoT}{Internet of Things}
\newacronym{a2a}{A2A}{Agent2Agent}
\newacronym{odrl}{ODRL}{Open Digital Rights Language}
\newacronym{rag}{RAG}{Retrieval Augmented Generation}
\newacronym{dsp}{DSP}{Dataspace Protocol}
\newacronym{kind}{KinD}{Kubernetes in Docker}
\newacronym{api}{API}{Application Programming Interface}
\newacronym{http}{HTTP}{Hypertext Transfer Protocol}
\newacronym{tc}{TC}{Test Case}
\newacronym{vc}{VC}{Verifiable Credential}
\newacronym{did}{DID}{Decentralized Identifier}
\newacronym{dcp}{DCP}{Decentralized Claims Protocol}
\usepackage{hyperref}
\usepackage{cleveref}      

\makeatletter
\def\ps@pprintTitle{%
  \let\@oddhead\@empty
  \let\@evenhead\@empty
  \let\@oddfoot\@empty
  \let\@evenfoot\@oddfoot}
\makeatother

\lstdefinestyle{jsonstyle}{
    basicstyle=\ttfamily\scriptsize,
    breaklines=true, breakatwhitespace=true,
    breakautoindent=true, breakindent=10pt,
    frame=single, rulecolor=\color{black!30},
    backgroundcolor=\color{gray!5},
    showstringspaces=false, columns=fullflexible,
    keepspaces=true, captionpos=b
}

\newcommand{\cmark}{\ding{51}}
\newcommand{\xmark}{\ding{55}}

\begin{document}
\begin{frontmatter}

\title{Agentic Orchestration of EDC Data Space Retrieval and Natural Language Driven Data Processing
}

\author[1]{Marla Grunewald\corref{cor1}}
\ead{marla.grunewald@tu-braunschweig.de}

\author[1]{Fin Gentzen}
\ead{f.gentzen@tu-braunschweig.de}

\author[1]{Iulisloi Zacarias}
\ead{i.zacarias@tu-braunschweig.de}

\author[1]{Admela Jukan}
\ead{a.jukan@tu-braunschweig.de}
  
\cortext[cor1]{Corresponding author}
\address{Institute of Computer and Network Engineering, Technische Universit\"at Braunschweig, Germany}

\begin{abstract}
Participating in a data space still requires the \gls{dsp} negotiation flow to be hand-coded against each provider's access policy. We present a multi-agent architecture that automates this process using natural language requests. Built on LangGraph over an \gls{edc}, the design separates concerns into cooperating agents. The orchestrators role is to route and plan. The data extraction agent's role is to expose the \gls{dsp} protocol as three composable tools (catalog discovery, contract negotiation and download). The agent also gates negotiations on the consumers' verifiable credentials, refusing assets rather than fabricating them. Furthermore, there are Downstream \gls{ml} and Visualization agents, as well as a final Verification agent, which confirm that the request was actually fulfilled and recheck the intended operation against the contracts usage policy. In a primary benchmark involving 600 runs, the system achieved 89 \% task success with zero policy violations when evaluating multi-dimensional policy reasoning with a locally hosted open-weight model. It failed safely through refusal rather than misrouting. The reliability of the system is largely dependent on the phrasing of requests: technical and expert requests succeed up to 98 \% of the time, whereas casual and under-specified requests succeed 72 \% of the time. Furthermore, in a secondary evaluation of 600 runs isolating hierarchical geographic access control, the framework successfully blocks 100 \% of unauthorized regional requests. While the verification agent hallucinates that 1\% of these requests, the system logs confirm that no data is ever extracted and the framework simultaneously achieves an end-to-end verified delivery rate of 86\% for eligible consumers. 
\end{abstract}


\end{frontmatter}

\section{Introduction}

As \gls{ai} applications become more ubiquitous, data remains one of the most valuable currencies. Worldwide, data centers are being built to manage and store the growing volume of data flowing for AI training and applications. Because the market is competitive and AI training requires large amounts of data across multiple domains, sharing data is a sensitive topic~\cite{vake2025open}. Traditionally, once data is exchanged, tracking its use or onward distribution becomes almost impossible. To solve this, the concept of common data spaces emerged in the 2010s with the founding of the \gls{idsa} from a worldwide alliance of companies and research institutions \cite{jarke2019data}. Data spaces facilitate secure, sovereign data exchange, a critical requirement for protecting personal privacy and preserving data integrity. Crucially, data spaces let data owners retain full control over the terms and volume of data they share with their counterparts~\cite{BACCO2024110969}. 

While frameworks like Gaia-X provide the infrastructure for sovereignty, they lack a user-centric interface that bridges the gap between high-level natural-language intent and low-level connector protocols.
Data exchange requests enter the data space via client software called a connector, which usually also handles user authentication. The connector facilitates communication between participants in a data space and ensures security and sovereignty \cite{BACCO2024110969}.
This communication is governed by the \gls{dsp}, which is a set of interoperability specifications defining how connectors discover datasets, negotiate usage contracts, and execute secure data transfers \cite{DSPProtocol}. In practice, the \gls{dsp} involves a complex, multi-step process. A consumer requests a catalog; the provider returns heavily structured metadata, and both parties must agree on a rigid policy framework (such as \gls{odrl}) using cryptography before any data can be transferred. Because these low-level \gls{dsp} negotiations require precise, formatted \gls{api} calls and extensive knowledge of data space mechanisms, manual participation is extremely challenging for non-experts, hindering broader use of sovereign data ~\cite{s26041193,RUELICKE2027108347}.

To address this bottleneck, Agentic AI systems offer a promising solution by abstracting complex software interactions behind natural-language interfaces. An agentic system is one in which an AI, such as a Large Language Model (LLM), can autonomously execute a sequence of actions in response to a single user request using a designated toolset. By breaking complex tasks into smaller subtasks and utilizing advanced reasoning and iterative planning, an agentic AI can navigate multi-step problems independently  \cite{10962241}. While the user describes their objective in natural language, the tooling behind the agent translates this description into a series of system calls, interacting with specific protocols on the user's behalf and abstracting the execution results back into an easily understandable output.

In this paper, we propose a novel architecture that leverages the agentic paradigm to let users operate data space connectors using natural language. We utilize a state machine-based agent orchestrator (LangGraph) to operate the \gls{edc} as a dynamic tool. 
Crucially, to avoid compromising the core principle of data sovereignty, the \gls{ai} agent must operate strictly within the policy boundaries enforced by the \gls{edc}. We further enhanced the system with tools designed to compile, train, and execute \gls{ml} models on the extracted data. 

The primary contribution of this paper is the introduction of a novel agentic data space orchestration framework.
A LangGraph-based architecture bridges high-level natural language intent with low-level \gls{dsp} negotiations across a multi-provider setup. A fundamental component of this contribution is the implementation of multi-dimensional policy reasoning directly within the agent's execution loop. The proposed system can autonomously parse and enforce complex \gls{odrl} policies across four distinct dimensions. Furthermore, this paper contributes a rigorous empirical evaluation of the proposed autonomous framework. We deploy a system architecture comprising three \gls{edc} providers and one consumer within a \gls{mvds} \cite{eclipse_mvd}. By leveraging the agentic \gls{ai} approach, we construct an end-to-end pipeline that autonomously pulls data from \gls{edc} nodes to directly train downstream machine learning models, using the MNIST dataset as a representative ML workload. Through comprehensive test cases and stylistic prompt variations, we demonstrate the system's ability to seamlessly perform cost-benefit optimization, aggregate federated data fragments, and recover from multi-node negotiation failures without human intervention. 

The remainder of this paper is structured as follows. Section \ref{sec:relwork} reviews related work, followed by our architectural design in Section \ref{sec:arch}, implementation details in Section \ref{sec:imp}, the empirical evaluation in Section \ref{sec:eval}, limitations in Section \ref{sec:limits}, and conclusion in Section \ref{sec:concl}.

\section{Background and Related Work}\label{sec:relwork}
\subsection{Data Space Connectors}
\begin{figure}[htb]
    \centering
    \includegraphics[width=\linewidth]{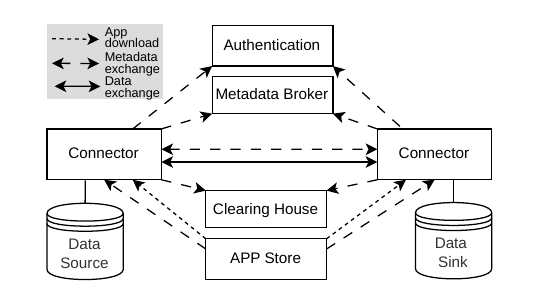}
    \caption{\gls{ids} reference Architecture for data space connectors as defined in \cite{IDSRef}.}
    \label{fig:idsRefArch}
\end{figure}

Data spaces are distributed systems where participants host their own data rather than relying on a central repository. To share and receive data, a data space connector is required. Figure \ref{fig:idsRefArch} illustrates the \gls{ids} reference architecture published by the \gls{idsa} \cite{IDSRef}.
The core entities in a data exchange are the data provider and data consumer. Before any exchange occurs, both participants must authenticate via an authority. The consumer can then utilize a metadata broker to discover available data offerings. Once a suitable connector is found, the two parties negotiate the terms of use. If successful, a clearing house logs the negotiations and subsequent data transfers. To further ensure privacy, specialized \gls{ids} apps can process, filter, or anonymize data as close to the source as possible (a data-to-code process) and are distributed via a secure \gls{ids} APP Store \cite{IDSRef}.

While initial access mechanisms secure entry points to these connectors, continuous data usage control is essential to safeguard data sovereignty after transfer~\cite{Jung2022}. To specify these rules, the \gls{ids} Information Model extends existing standard vocabularies like the \gls{odrl} to bridge the gap between descriptive policies and technical runtime enforcement across federated systems~\cite{10.1007/978-3-030-62466-8_12}.  As \gls{odrl} has emerged as the \textit{de facto} standard for modeling usage preferences in European dataspaces, ensuring consistent, unambiguous evaluation of these policies has led to the development of formally grounded, high-performance engines like the OVAL evaluator, which support both static access control and real-time streaming monitoring~\cite{salas2026formallygroundedodrlevaluator}. However, manually configuring and managing these machine-readable policies remains a significant barrier to broader adoption.

The objective of our work is to ensure that the data space connector is accessible to non-expert users while complying with the complex nature of data usage control. To accomplish this, we engineered a natural language-driven agentic system that interfaces with a data space using the \gls{edc}.

\subsection{Agentic and LLM-based Data Interactions}
The integration of \gls{llm} and agentic systems has fundamentally transformed data interactions, shifting from manual processing to dynamic, natural-language-driven orchestration. Multi-agent frameworks can now autonomously execute end-to-end data science pipelines, decomposing complex tasks such as data cleaning, feature engineering, and model evaluation through specialized roles and external tool invocation~\cite{11198037}. To further optimize these workflows and reduce the overhead of data movement, recent advancements have embedded LLM capabilities directly into relational database management systems. Frameworks like FlockMTL allow users to seamlessly execute semantic analysis and \gls{rag} natively within SQL, minimizing complex system orchestration~\cite{10.14778/3750601.3750685}. 
Extending these capabilities to data spaces requires standardized, secure interfaces that limit both credential misuse and the effects of \gls{llm} hallucinations. The authors in ~\cite{s26041193} address this by using the \gls{mcp} to connect LLMs to an IoT data space through the standardized ETSI NGSI-LD API. Their MCP server exposes read-only tools for type discovery and entity retrieval, so the LLM can answer natural-language queries over contextual data. The server validates tool invocations to catch hallucinated types or attributes, and it enforces per-user authorization using DPoP-bound access tokens, so a token leaked through the MCP host cannot be reused.
As automated agents increasingly navigate federated data ecosystems, robust infrastructure is required to govern their data discovery and consumption. The \gls{edc} connector provides this foundational mechanism by automating interface discovery, metadata brokering, and contract negotiation, which drastically accelerates the integration of distributed assets, as demonstrated in decentralized energy management systems~\cite{11357554}. Furthermore, the \gls{edc} connector ensures data sovereignty in collaborative, privacy-preserving AI applications, successfully enforcing usage policies during Federated Learning across sensitive, multi-institutional healthcare environments~\cite{10826122}. 
\begin{figure*}[!ht]
    \centering
    \includegraphics[width=1\linewidth]{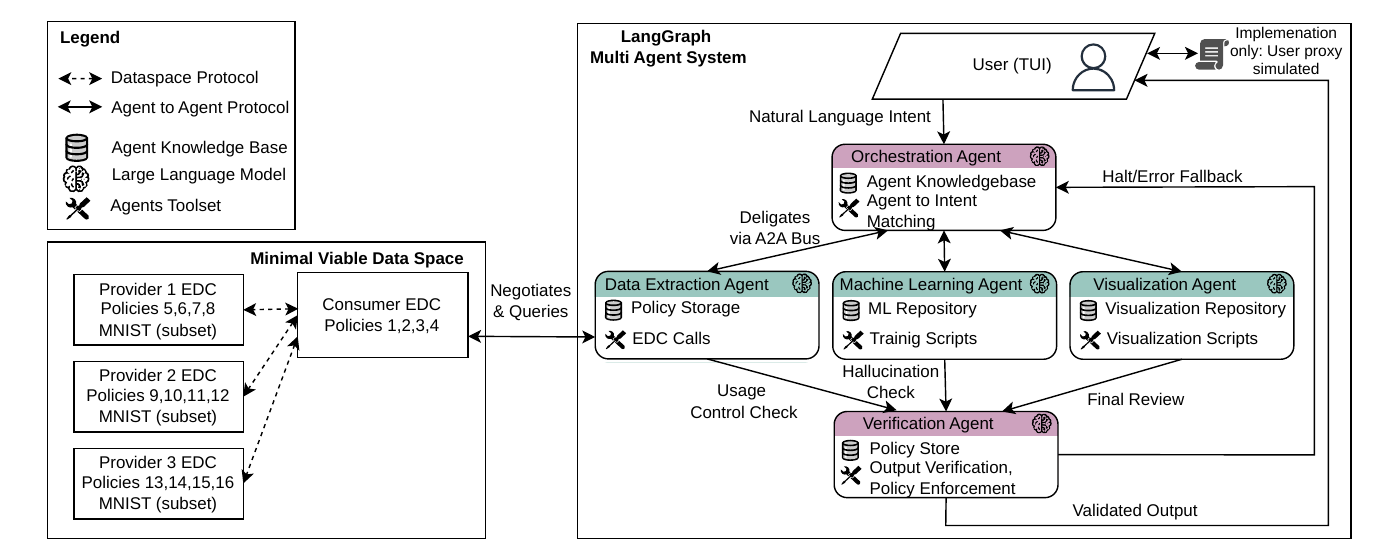}
    \caption{The proposed system architecture consists of the LangGraph Agentic System and the \gls{mvds}. While the purple agents are mandatory for every call, the system requires at least one, and can utilize up to all three, of the green worker agents. In order to test the setup, we have added a file with user responses in order to automate the testing of the setup, this file is visible next to the TUI in the architecture.}
    \label{fig:Arch}
\end{figure*}
Despite these advancements, a critical gap remains at the intersection of autonomous AI and sovereign data exchange. As highlighted in a recent \gls{idsa} position paper ~\cite{turkmayali2026dataspaces}, standard AI pipelines lack a built-in compliance and authorization required to cross organizational boundaries. While data spaces inherently solve these coordination challenges, such as identify verification, trust and provenance, they were primarily designed for static software interactions.  
Deploying autonomous agents introduces novel architectural demands. While credential issuers and users may supply static identities and constraints upfront, agents require dynamic, real-time mechanisms to parse foreign \gls{odrl} policies, evaluate trade-offs and execute machine to machine contract negotiations. 
Without these dynamic mechanisms, an agent risks either stalling during data acquisition or more critically, hallucinating compliance and inadvertently violating strict geographic and monetary constraints. 
Our work directly addresses this call to action by proposing a framework were an agentic system dynamical parses and negotiates complex \gls{odrl} usage policies, successfully bridging high-level AI orchestration with robust data space governance to enable fully automated, policy compliant machine learning pipelines.

\section{System Architecture} \label{sec:arch}

\subsection{Dataspace Infrastructure and Topology} \label{subsec:dataspaceinfrastructure}
The simulated dataspace environment consists of a \gls{mvds}, which comprises four distinct, logically isolated network nodes. Each node operates an \gls{edc} connector, dividing operations between a control plane for \gls{dsp} negotiations and a data plane for physical data transfer. Three of these nodes act as external providers, hosting target datasets such as MNIST data. The control plane of each provider enforces strict \gls{odrl} usage policies across dimensions such as geography, computational cost and allowable entity types. The fourth node acts as the consumer gateway securely connecting the local environments to the \gls{mvds}. It is the sole gateway for local agents, managing cryptographic identity and executing transfers without hosting any initial data. 
Trust within the \gls{mvds} is established through a dedicated issuer service, which acts as the dataspace's trust anchor. Following the \gls{dcp}, each participant is identified by a \gls{did} and holds \glspl{vc} in its IdentityHub. The issuer service signs credentials attesting to participant attributes that are relevant to the providers' \gls{odrl} policies, such as a \emph{LocationCredential} certifying the consumer's jurisdiction and a credential certifying its organizational entity type. During catalog requests and contract negotiation, the consumer presents these credentials as verifiable presentations, and the provider's control plane verifies them before evaluating its usage policies. Policy constraints are therefore enforced cryptographically at the protocol level, independently of the agents' reasoning.

\subsection{Multi-Agent Orchestration Engine}
\label{subsec:multiagent engine}

The AI environment features an orchestration engine that leverages LangGraph to bypass traditional linear execution scripts. Instead, a dynamic \gls{a2a} communication bus is utilized. 
The Framework consists of five specialized agents that operate collaboratively on a shared execution state: an Orchestrator, a Verification Agent, and three worker agents dedicated to extraction, \gls{ml}, and visualization. The orchestrator acts as the central routing node, communicating with the user via a \gls{tui} and managing all inter-agent communications. When a natural language prompt is received, it queries a centralized knowledge base to perform agent-to-intent matching. This allows the Orchestrator to dynamically determine which subset of agents is necessary for a specific task. For instance, a simple data retrieval request bypasses the downstream machine learning agent entirely. 

Once invoked, tasks are delegated to specialized worker agents. All worker agents share a common internal architecture consisting of a localized knowledge base for state tracking, a specialized tool set, and an internal verification block that validates task execution against predefined limits. Crucially, the Data Extraction agent serves as the core engine for autonomous contract negotiation. Rather than executing hard-coded rules, this agent dynamically parses machine-readable \gls{odrl} policies retrieved from federated catalogs.
It maps the users' localization profile, such as geographic location, available budget, and organizational entity, against the providers' constraints across four distinct policy dimensions.
Using the \glspl{llm} advanced reasoning capabilities, the agent evaluates complex cost-benefit trade-offs and autonomously selects the optimal, policy-compliant provider. Once a decision is made, the agent triggers the consumer \gls{edc} to execute the \gls{dsp} handshake, fully automating contract negotiation and data aggregation and eliminating the need for human intervention. Responsible for downstream data processing, the Machine Learning Agent maintains an internal repository and actively monitors training parameters to ensure optimal model compilation and evaluation on the extracted data. Finally, the Visualization Agent formats the final user output, such as tables, graphs, or images, using Python-based tools aligned with the requested format. 

\subsection{Verification and Policy Compliance} 
\label{subsec:policy}
As the system interacts with sovereign data space boundaries by design, the Verification Agent acts as a mandatory security and compliance gateway. Any interaction involving data exchange requires its involvement to perform two critical functions. Firstly, it acts as an independent auditing mechanism for the autonomous negotiations. As \glspl{llm} can be susceptible to hallucinations, the Verification Agent strictly validates the data extraction agents' proposed contract against the provider's original \gls{odrl} constraints prior to the final data transfer. For example, it ensures that the AI has not falsely assumed geographic compliance or ignored a monetary limit. Upon detecting any policy deviation, the Verification Agent immediately halts the system operation, triggers an error-handling fallback, and notifies the user. Secondly, it performs final output validation. Before passing the generated models or visualizations back to the Orchestrator for \gls{tui} rendering, the agent reviews the worker output to identify and mitigate potential AI hallucinations. This decoupled, graph-based architecture ensures high extensibility, enabling the seamless integration of heterogeneous agents while maintaining rigorous data sovereignty.

\section{Implementation} 
\label{sec:imp}

\subsection{Software Environment and System Setup}

The system is deployed locally within a containerized environment to ensure reproducible evaluation while maintaining strict logical network isolation. The \gls{mvds} is built on the \gls{edc} framework and is orchestrated using \gls{kind}. Within this local Kubernetes cluster, the infrastructure is distributed across multiple distinct pods. Among the services required for the data space, one consumer and three providers are launched. The \gls{edc} pods are configured to distinctively divide the control and data planes and all communication between participants adheres strictly to the official \gls{dsp}.
The agentic environment is developed using LangGraph to manage shared execution states and inter agent communication. The framework uses a locally deployed Mistral (Mistralai/Magistral-Small-2509) model to power the cognitive reasoning of the agents. Executing the \gls{llm} entirely on-premises rather than relying on external commercial \glspl{api} is a critical architectural decision. This guarantees that sensitives organizational constrains embedded within \gls{odrl} polices, as well as private data space metadata and extracted datasets, are never exposed to third-party cloud providers.  This strict containment aligns with the principle of data sovereignty and privacy principles required for secure data space participation. Finally, user interaction is facilitated through a \gls{tui} constructed with the textual framework, providing real-time visual rendering of agent states, active \gls{odrl} policy parsing and contract negotiation status. 

\subsection{ODRL Policy Representation and EDC Integration}

To enable dynamic policy reasoning, the Data Extraction agent communicates with the consumer \gls{edc} via a custom REST \gls{api} tool layer. During the discovery phase, the agent uses the consumer to retrieve all providers catalogs. Fetching these catalogs requires a baseline membership policy, which the consumer possesses and automatically presents. To simulate a realistic, noisy data ecosystem, a variety of heterogeneous assets (totaling 30 assets across the \gls{mvds}), were added to each providers catalog. Distractor datasets (e.g. irrelevant image or tabular data) and policy conflict traps are intentionally included in these catalogs. The latter are assets which contain the correct target data, but are restricted by conflicting geographic or monetary constrains. This setup rigorously tests the agents ability to semantically filter catalogs and perform multi dimensional trade-off reasoning without human intervention. 

The permissions, prohibitions, and duties embedded in these catalogs are extracted and transformed into a structured JSON payload. As outlined in Table \ref{tab:odrl_policies}, the extraction process targets 4 distinct evaluation dimensions. Firstly, geographic constraints evaluate spatial location properties, requiring the agent to understand hierarchical boundaries, for example, distinguishing between Lower Saxony in Germany and the wider European Union. Secondly, monetary limits compare the requested payment amount with the user's allocated budget, which can range from free access to premium tiers (Provider 3). Thirdly, usage constraints verify the permitted usage purpose, forcing agents to differentiate between tasks such as machine learning training, pure inference, and data visualization. Finally, the entity profile dimension cross-references assignee roles and evaluates whether the user will publish the results openly or semi-openly, or use them for personal gain.

\begin{table*}[htbp]
\centering
\caption{Simulated ODRL Policy Constraints Across MVD Providers}
\label{tab:odrl_policies}
\begin{tabular}{@{}lcccrccccccc@{}}
\toprule
   & \multicolumn{9}{c}{\textbf{Policy Dimension}}\\ 
   \cmidrule(lr){2-12}
   & \multicolumn{3}{c}{Geographic} & & \multicolumn{3}{c}{Allowed Usage} & \multicolumn{3}{c}{Entity Profile} &  \\
   \cmidrule(lr){2-4} \cmidrule(lr){6-8} \cmidrule(lr){9-11}
\textbf{Provider} & EU & Germany & Lower Saxony & Monetary & T & I & V & PP & RI & CO & (Closed)\\
   \midrule
\textbf{Provier 1 ($P_1$)} & \cmark & \cmark & \cmark & € 0.00    & \cmark & \cmark & \cmark & \cmark & \cmark & & \cmark \\
\textbf{Provier 2 ($P_2$)} &        & \cmark & \cmark & € 500.00  &        & \cmark & \cmark & & \cmark & \cmark & \cmark\\
\textbf{Provier 3 ($P_3$)} &        &        & \cmark & € 2000.00 & & & \cmark & \cmark & \cmark & \cmark & \xmark \\
\bottomrule
\multicolumn{12}{l}{\footnotesize Allowed Usage: Training (T), Inference (I), Visualization (V)} \\
\multicolumn{12}{l}{\footnotesize Entity Profile: Private Person (PP), Research Institute (RI), Commercial (CO)}
\end{tabular}
\end{table*}

The language model navigates the simulated traps by mapping the user's localized profile against these specific constraints. Once the agent has selected a legally valid target asset based on this multidimensional matrix, it sends a request to the consumer \gls{edc} control plane management \gls{api}. The \gls{edc} then handles the asynchronous \gls{dsp} handshake. Once the parties reach a final agreement, the data plane issues an HTTP transfer request, streaming the requested dataset directly into the local workspace memory.

\section{Evaluation}
\label{sec:eval}

\subsection{Benchmark Design and Evaluation Methodology}

To empirically evaluate the framework while maintaining the strict isolation of variables, two distinct experimental runs were conducted as part of the benchmark design. Running two separate evaluation tracks, one focused on multidimensional policy reasoning and system performance, and the other dedicated strictly to geographic constraints, ensures that distinct capabilities can be assessed individually. This separation simplifies the evaluation process and prevents errors from different sources becoming conjoined, which would be difficult to identify if all possible scenarios were combined into a single monolithic run. The first experimental run focused on policy constraints and system performance. The benchmark systematically cross-tested five test cases (TC1–TC5), four agent activation levels (L1–L4) and three prompt phrasing styles. This yielded a total of 60 distinct configurations. The five test cases evaluate dynamic policy reasoning under different data space conditions. Three of these cases involve exclusive matches to a single provider (P1, P2 or P3), one presents a multi-provider scenario in which the agent must optimize for the cheapest qualifying offer and the last represents an unsatisfiable request that requires the agent to autonomously refuse negotiations. Execution complexity is scaled across four agent activation levels ranging from isolated data extraction, to extraction paired with either machine learning or visualization tasks, to full orchestration, which exercises all worker agents concurrently. To evaluate the robustness of the agents' natural language understanding, queries are formulated using casual, technical and expert phrasing styles. This methodology tests the agents' ability to accurately extract implicit constraints or proactively request missing information, rather than making unverified inferences.

Conversely, the second experimental run isolates hierarchical geographic access control by systematically varying consumer location attributes, ranging from highly specific local jurisdictions to out-of-bounds international locations, against providers with progressively stricter spatial policies. In this secondary track, all non-geographic credentials are pre-satisfied to ensure negotiation and routing outcomes are solely determined by geographic boundary enforcement.

To maintain strict experimental isolation, the corresponding consumer identify and verifiable credentials are dynamically provisioned prior to each run and cleanly restored upon completion. 
Each evaluation run is executed end-to-end and monitored in real time via the orchestration engines streaming interface. From a single pass, two synchronized telemetry artifacts are derived: a structured (CSV) log capturing the precise timing and execution sequence of the active agents and an audit transcript detailing every tool call, dynamic routing decision and formal verification verdict (JSONL). A run is classified as successful only if its observed execution trace aligns with the established ground truth for that scenario. For admissible request, success requires negotiating with the correct provider, receiving explicit clearance from the Verification Agent and completing all the requested downstream processing steps. For unsatisfiable scenarios, success is defined as correct autonomous termination that preserves sovereign boundaries and yields zero data leakage. 

\subsection{Policy Constraint and System Performance Evaluation}

Because the LLM-driven agent is non-deterministic, we evaluated a full factorial of 60 configurations (5 test cases × 4 autonomy levels × 3 prompt styles) and repeated each 10 times (600 runs total). Repetition is essential: 23 of the 60 configurations yielded mixed outcomes across repetitions. The ten repetitions let us estimate a per-configuration success probability (95\% Wilson CI ±17.5 pp on average), which is what exposes the style- and level-dependent failure modes, while pooling all 600 runs yields an overall task-success rate of 88.7\% (95\% CI [85.9\%, 91.0\%]).l

\begin{figure}
    \centering
    \includegraphics[width=\linewidth]{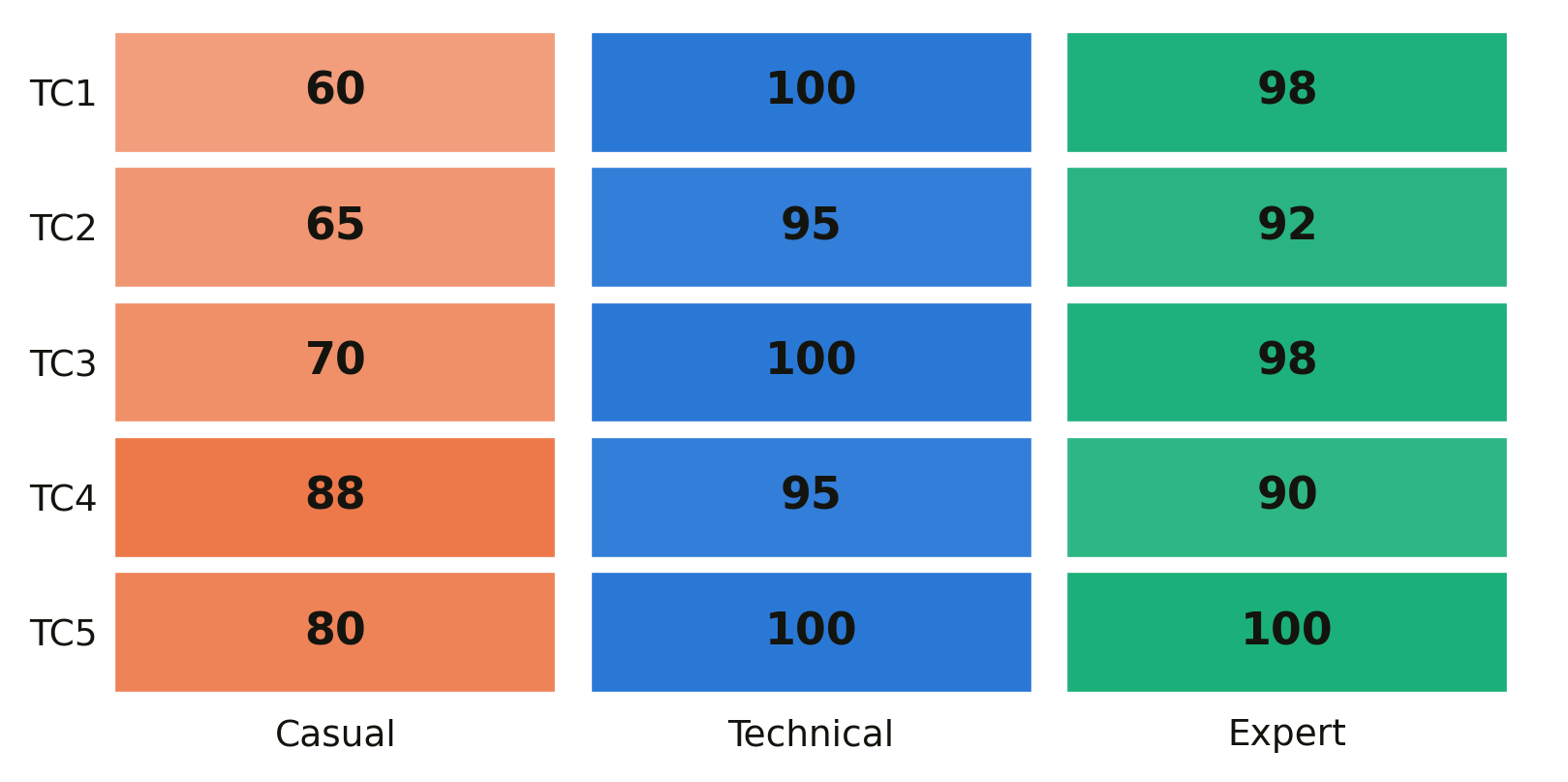}
    \caption{The heatmap shows the success rate of the \gls{tc} by Prompting style in \%. }
    \label{fig:heatmapStylesHeatmap}
\end{figure}
The framework's ability to extract constraints and negotiate effectively was heavily influenced by the semantic structure of the user queries.
As illustrated Fig \ref{fig:heatmapStylesHeatmap}, success rates varied significantly across prompting styles. Technical and expert phrasing yielded the highest robustness, achieving success rates of 98\% and 96\%, respectively. In contrast, casual conversation prompts performed the least well, dropping the success rate to 72\% across all test cases. This performance gap suggests that while the Mistral model is highly capable of multidimensional policy reasoning, its performance degrades when implicit constraints must be inferred from ambiguous or colloquial language, rather than explicitly stated parameters.

\begin{lstlisting}[style=jsonstyle, caption={Example of the used prompts.}, label={lst:prompts}]
{
  "level": "L4_extract_ml_vis",
  "casual": {
    "prompt": "Get me those digit pics, train a model, and show me how it did?",
    "user_answer": "Free please, just train something small and then show me how the training went."
  },
  "technical": {
    "prompt": "Download MNIST (budget 0, ml_use training), train a CNN (2 conv layers, 2 epochs), then plot the training accuracy/loss curves.",
    "user_answer": "Budget 0, ml_use training."
  },
  "expert": {
    "prompt": "Obtain the MNIST asset (budget credential 0, ml_use training), train a 2-conv CNN for 2 epochs, and visualize the training-history curves from the saved history JSON.",
    "user_answer": "Budget credential 0, ml_use training."
  }
}
\end{lstlisting}

\begin{figure}
    \centering
    \includegraphics[width=1\linewidth]{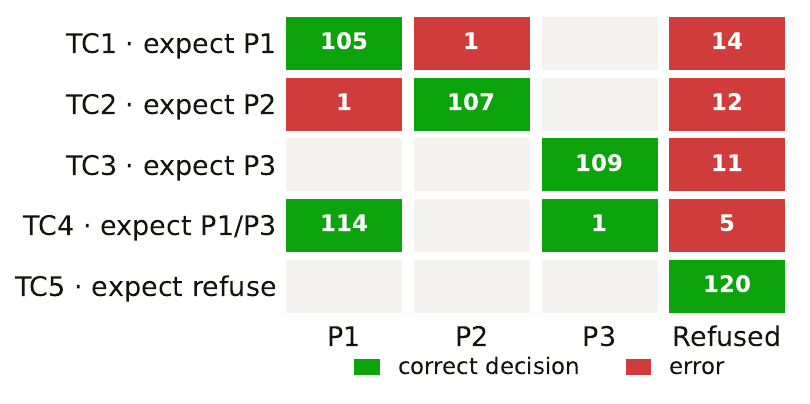}
    \caption{The Figure shows the correct and incorrect routing decisions for the Data Extraction Agent. Green cells are correct decisions, red cells are errors.}
    \label{fig:resultsprovider}
\end{figure}

Going beyond basic pass/fail metrics, Figure 4 uses a confusion matrix to illustrate the specific negotiation and routing decisions made by the agents. Green cells represent correct routing, while red cells denote errors. Crucially, the red cells highlight a conservative failure mode rather than policy violations.
Almost all errors involve the system over-refusing legitimate requests, prioritizing safety over access. Within this first experimental track, true misrouting is exceptionally rare, and Test Case 5 confirms zero instances of data leakage across all 120 unsatisfiable requests. A notable anomaly occurred in Test Case 4 (TC4), designed to evaluate cost optimization. Here, assets from both Provider 1 (P1) and Provider 3 (P3) met all legal and user constraints. While P1 was significantly cheaper, the agent routed to P3 in one instance. Though this represents a minor failure in economic optimization, the decision remained strictly compliant with all access policies, further demonstrating the system's strong adherence to legal validity.

\begin{figure}
    \centering
    \includegraphics[width=1\linewidth]{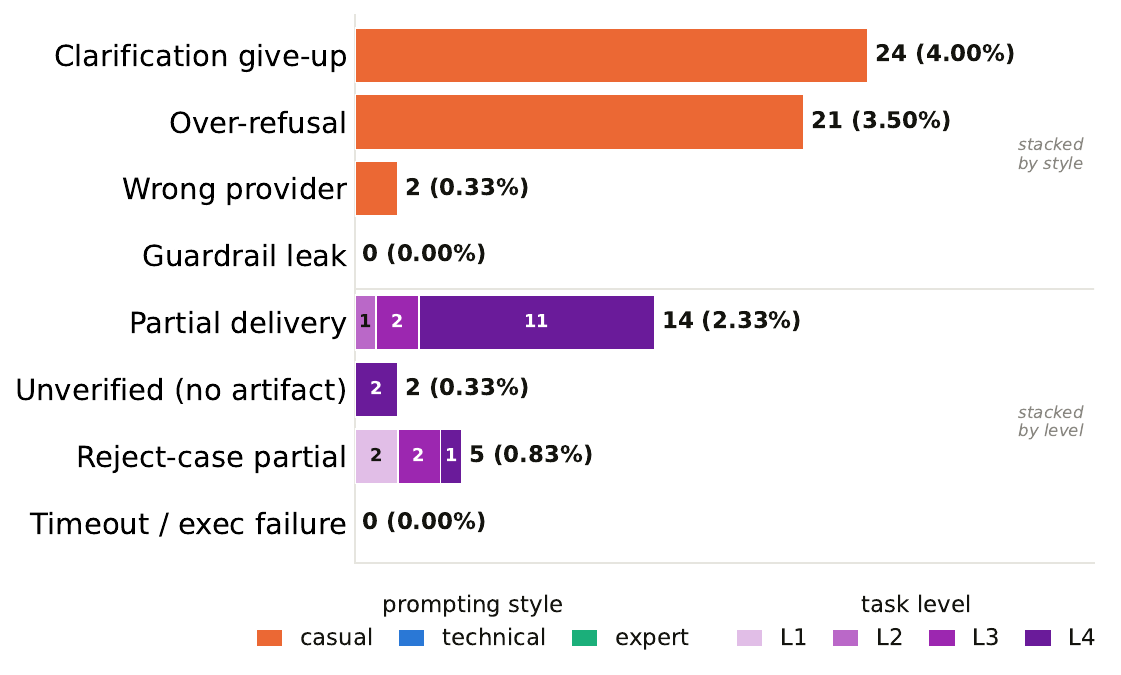}
    \caption{Failure analysis, where task failed and how often. The overall number of tasks is 600 runs.}
    \label{fig:failures}
\end{figure}

Figure \ref{fig:failures} gives a detailed overview of the failures depicted in Figure \ref{fig:resultsprovider}. The discrepancy in total error counts observed between Figures \ref{fig:resultsprovider} and \ref{fig:failures} is due to the fact that Figure \ref{fig:resultsprovider} exclusively evaluates the agent's individual routing decisions, whereas Figure \ref{fig:failures} reflects the system's overall execution health.

The experimental error runs fall into two main categories. The first category comprises four error types associated with prompting style, which only occurred under the \textit{casual} writing style condition. This phenomenon is largely attributable to the automated testing methodology employed. Specifically, the simulation relied on prescripted fallback prompts to emulate user replies. When these hardcoded replies lacked sufficient detail or accuracy, the conversational agent prematurely terminated the dialogue, with these outcomes being classified as either \textit{clarification give-up} or \textit{over-refusal}. While a human operator would likely have provided the necessary context to sustain the interaction, the static nature of the automated replies forced the agent to abort. Although this methodology yields a high error rate on the surface, it highlights a robust system safety mechanism: the agent resists hallucination, choosing to terminate the process safely rather than providing an incorrect response.
Furthermore, incorrect provider routing was observed in Test Cases 1 and 2, as depicted in the upper-left quadrant of Fig. \ref{fig:resultsprovider}. Crucially, no guardrail violations were recorded across any of the 600 runs in the primary multi-dimensional benchmark, ensuring that no unauthorized data was exposed to the user under standard policy configurations.

The second category of errors in Fig. \ref{fig:failures} is categorized by agent level and specifically characterizes failures that occurred during interaction with the verification agent. In these instances, execution runs were completed but subsequently faced full or partial rejection during the verification phase. For instance, of the 14 cases of \textit{partially delivered} tasks observed, the system successfully routed to the correct provider and generated the necessary data file, but the verification agent failed to validate the run upon completion.  The \textit{unverified} error category displays similar behavior, with the key difference that no data artefacts were produced. The \textit{partial rejection} case was exclusively observed in the unsatisfiable TC5, where the system neither served the dataset nor executed a clean refusal, but importantly data security remained completely uncompromised. Finally, the evaluation revealed no instances of general system failures due to timeouts or other system-level exceptions.

\begin{figure}
    \centering
    \includegraphics[width=1\linewidth]{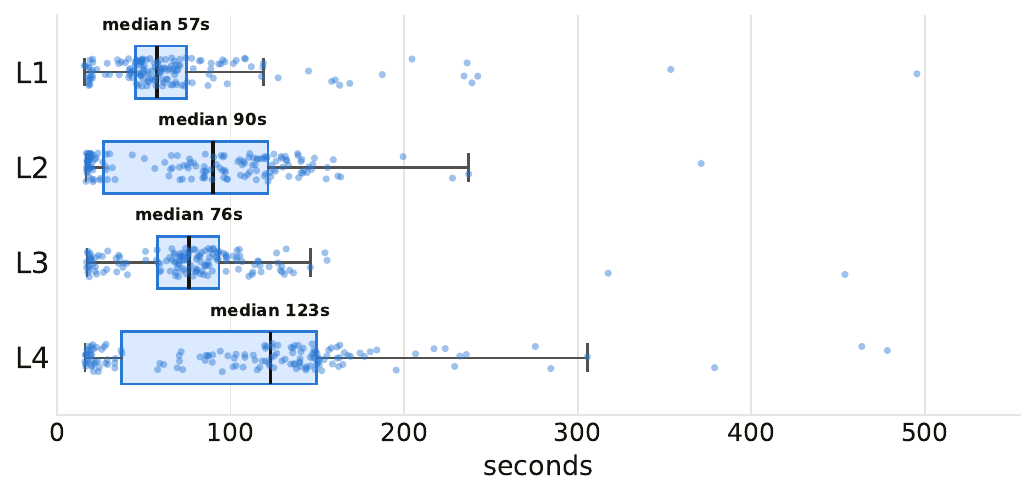}
    \caption{The figure shows the latency by Agent Level in seconds in a boxplot where the whiskers are located at 1.5xIQR. }
    \label{fig:latency}
\end{figure}

Figure \ref{fig:latency} presents a horizontal box plot detailing the execution latency across four distinct task levels (L1–L4), with individual runs overlaid as faint scatter points. The results demonstrate that latency scales proportionally with task complexity. The baseline task, L1 (data extraction only), exhibits the lowest median latency at 57 s. When the pipeline is expanded to include data visualization (L3), the median latency increases to 76 s. In contrast, incorporating machine learning training into the extraction process (L2) results in a higher median latency of 90 s. Finally, L4 demonstrates the highest overall median latency at 123 s, reflecting the computational load of the complete end-to-end pipeline, which comprises data extraction, Convolutional Neural Network training, and results visualization.

\subsection{Geographic evaluation} \label{subsec:geo_results}
\begin{figure*}[htbp]
    \centering
    \includegraphics[width=0.85\linewidth]{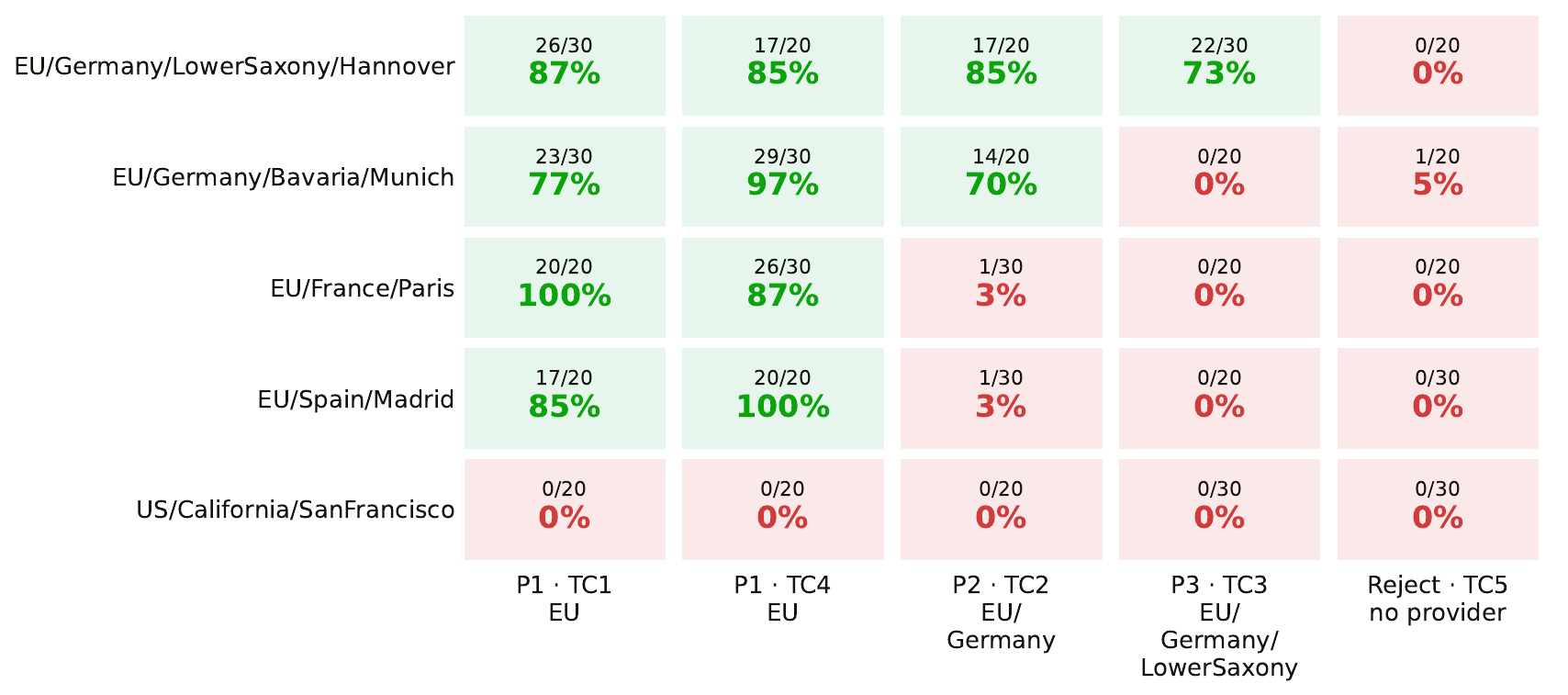}
    \caption{Gating Matrix of the tested geographic regions and Provider policies. Consumer locations are assigned in a 5-step round-robin cycle across the 600 runs (120 runs per location). Because each test case spans 12 consecutive runs (4 levels $\times$3 styles) and 12 is not divisible by 5, each block allocates 3 runs to two locations and 2 runs to the other three, shifting by $12 \bmod 5 = 2$ positions per test case. Since every 60-run replication restarts at the same phase, this split scales directly across 10 repetitions into the 20- and 30-run cell counts observed in the matrix.}
    \label{fig:geooptions}
\end{figure*}

The second experimental run specifically isolates geographic variables to assess boundary enforcement independent of other policy parameters.
Figure \ref{fig:geooptions} demonstrates how hierarchical geographic access control works end to end across the live data space. The five rows represent consumer locations, ordered from the closest match at the top (EU/Germany/Lower Saxony/Hannover) to the furthest match at the bottom (US/California/San Francisco). The five columns represent providers test case scenarios, whose policies require progressively stricter regions (EU, then EU/Germany, then EU/Germany/Lower Saxony). A green box indicates that the contract negotiation should be accepted, and a red box indicates that it should be rejected. Because every non-geographic credential (budget, classification, ML use) is pre-satisfied, geography is the only variable in this test run that can cause a rejection. The result is a clean staircase: Hannover sits inside all three required regions and is accepted everywhere, Munich fails only the strictest Lower Saxony policy, Paris and Madrid matches only the broadest EU policy and San Francisco matches nothing and is refused across the board. The cells show how many of the test cases where accepted and denied based on the geographic experiment. The figure shows that policy region requirements correctly and predictably grant access as location specificity varies.

\begin{figure*}[h]
    \centering
    \includegraphics[width=0.85\linewidth]{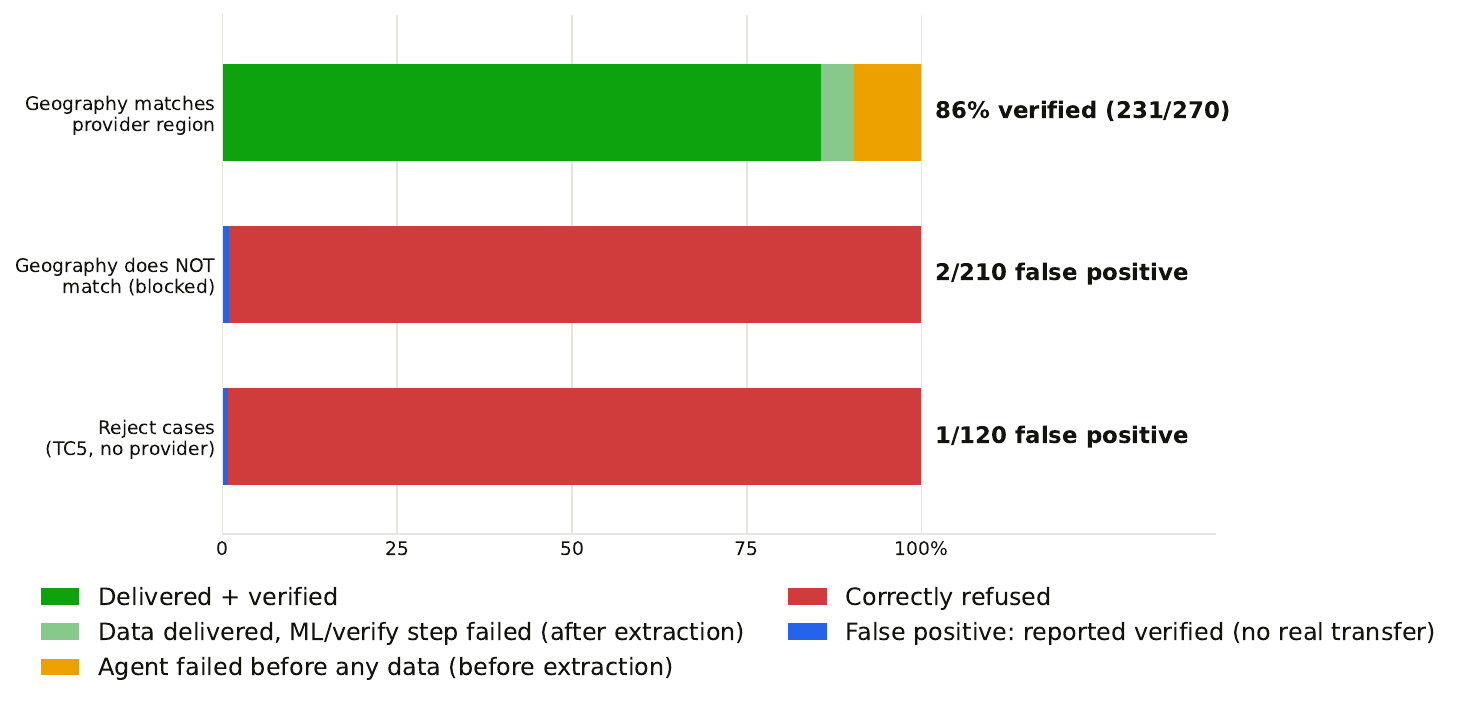}
    \caption{Results of executions regarding geographic policy satisfaction or blocking. The top bar represents requests where the consumer location matches the provider restrictions. The middle bar represents requests where the consumer location does not match the provider restrictions; therefore, the multi-agent system should stop the negotiation process. The bottom bar represents requests with no eligible provider; thus, the request should be ignored or blocked.}
    \label{fig:results_geography_run}
\end{figure*}

Figure \ref{fig:geooptions}  shows the pass/fail rates for each location-to-test-case pair in detail, while Figure \ref{fig:results_geography_run} groups these 25 cells from all 600 runs into three high-level compliance categories based on whether the consumer's location met the target policy. The top bar pools all 11 green cells where the geography matches the provider's region (n = 270), the middle bar pools the nine red cells across TC1–TC4 where the geography does not match and access must be blocked (n = 210) and the bottom bar pools the five cells in the final column for TC5 where no valid provider exists, regardless of location (n = 120). The reject bar corresponds to all runs in TC5, since TC5 has no valid provider and is therefore rejected regardless of location. The remaining runs are divided into allowed and blocked, and this split is uneven because the test cases require regions of different strictness: a broad requirement like the EU puts more eligible locations inside the region, while a strict requirement puts more eligible locations outside. The absolute ratios of each case are printed on the outside of the bar.
\par The top bar shows the runs where the geography matches the provider's required region. 86\% of these runs were completed as 'delivered+verified', 4.8 \% of the runs were marked as 'delivered+pipeline failure' and 9.6\% of the runs were marked as 'data extraction failure'. Therefore, none of these failures are policy leaks. The middle bar shows the runs where the geography does not match and access should be blocked. 
In these cases, the EDC blocks the process at the beginning of negotiations, resulting in 100 \% denial. Although the verification agent claimed that 1 \% (two out of 210) slipped through, system logs confirm that no contracts were made with the providers and no data was obtained. Similarly, the bottom bar, which represents TC5 reject cases where no eligible provider exists, shows that 100 \% of cases are correctly rejected, despite a hallucinated pass rate of 1 \% (one out of 120). Unlike the primary benchmark, where geographic regions remained static and no violations occurred, stress testing fine-grained, hierarchical spatial boundaries introduced rare false positives in the evaluation layer (three out of a total of 330 blocked runs), where the LLM-based verification agent incorrectly identified regional containment. As the logs confirm that the EDC intervened early, there were zero true violations.
 
The main difference is between the top bar, where access is usually allowed, and the two bottom bars, where access is always refused. The verified percentages measure full success: not just that a contract was signed, but also that data was delivered and verified. Overall, the figure shows that the geographic policy is permissive when appropriate, but also ensures that access is blocked when necessary. Any residual failures on the matching row are execution irregularities, while the rare anomalies on the blocked rows are strictly verification hallucinations rather than true security failures.

\section{Limitations}
\label{sec:limits}

While our proposed architecture and implementation are as generic and model-agnostic as possible and can work with most LLMs, we acknowledge that we tested our prototype on a small, lightweight LLM, and results may vary with mid- to large-sized LLMs. Other model families and different quantization levels may shift success rates and sensitivity to input prompts. Integrating a different model should be easy and smooth, since we use an intermediate HTTP API that lets us connect the developed multi-agent system to multiple models. We would also like to emphasize that our model choice strengthens our proposed solution by allowing users to self-host the LLM and keep data under their control.

Similarly, our three data providers host well-known, open datasets that we use to mock a real application, and the data may not be as heterogeneous as in real-world operational scenarios. Nevertheless, we expect most of our results to hold even with different datasets, since our algorithms consider metadata and data policies rather than the raw data stored by the providers. This decision also enables easy reproducibility, since data can be openly and easily found on the web.

We recognize that our human-in-the-loop responses to LLM clarification requests were automated using scripted input, and this may not reflect real human interaction. Therefore, the failures observed in the casual prompting experiment may partly stem from replaying the static file containing the supposed user's answer. Therefore, our results do not show how real non-expert users formulate requests or answers to the multi-agent system, since we have not validated the system in a user study.

It is important to emphasize that dataset usage after the negotiation phase is deliberately not a concern for our system. The agentic layers evaluate offers from the EDC control plane and select the best-suited offer based on our defined policies. We understand this is crucial for data sovereignty, but we would also like to stress that it is a complex task that requires strict control of data access, requiring solutions such as trusted execution environments. Also, the policies we used in our experiments cover four dimensions, accepting single values (See Table \ref{tab:odrl_policies}). We do not test logical compositions of constraints, temporal conditions, or conflicting permissions.

The verification agent is LLM-based and therefore probabilistic, it provides no formal guarantee of policy compliance. As reported in Section~\ref{subsec:geo_results} (Fig.~\ref{fig:results_geography_run}), the agent incorrectly approved roughly 1\% of contracts that violated geographic constraints, by hallucinating during the verification process. None of these resulted in a data transfer, because the provider's control plane rejected the negotiation after verifying the consumer's credentials against its \gls{odrl} policy. The verification agent's approval is therefore not sufficient on its own to
confirm that a contract is compliant. Moreover, because the verification agent uses the same model as the agents it audits, their errors may be correlated. Additionally, the verification agent only checks whether training and visualization tasks were completed, not their quality, which depends on the dataset, usage scenario, and application and is outside the scope of this work.

Finally, one of our assumptions is that the system will serve honest users on both ends (clients and providers). Our system consumes the catalog metadata, description, and policies as text and forwards them to the LLM, creating a potential channel for prompt injection. Such limitations are acceptable in a research prototype, but in a real-world scenario, such data should be encoded using an ontology and pass verification before feeding it to the LLM models. 

\color{black} 

\section{Conclusion} 
\label{sec:concl}

In this paper, we proposed a multi-agent framework based on LangGraph that bridges the gap between natural language intent and complex \gls{dsp} negotiations. Built to use the \gls{edc} as a tool, the system uses specialized agents operating on a locally deployed Mistral model to autonomously extract data and perform machine learning tasks while strictly preserving data sovereignty. 
Our initial 600-run benchmark test showed an overall success rate of 89 \%, demonstrating multi-node contract negotiation and cost-benefit optimization with no policy violations. A secondary evaluation of 600 runs, isolating hierarchical geographic policies, confirmed a 100 \% blocking rate for unauthorized regions, alongside an 86 \% verified delivery rate for compliant requests. System performance depended heavily on prompt phrasing across our evaluations: technical requests achieved a success rate of up to 98 \%, while casual, under-specified prompts dropped to 72 \%, failing safely through refusal rather than hallucinating compliance. End-to-end execution latency increased predictably with task complexity, peaking at a median of 123 seconds for the entire process. Future work will focus on integrating multi-turn clarification loops to resolve ambiguous intent, optimizing localized inference latency and evaluating the framework on broader, more heterogeneous datasets. Finally, to eliminate the rare 1 \% false acceptance rate observed during complex geographic stress testing, the vulnerabilities of the LLM-based verification agent must be addressed by integrating a deterministic, formal policy evaluation engine to enforce strict security.


\bibliographystyle{elsarticle-num}
\bibliography{references.bib}

\end{document}